\def\year{2018}\relax
\documentclass[letterpaper]{article} %DO NOT CHANGE THIS
\usepackage{aaai18}  %Required
\usepackage{times}  %Required
\usepackage{helvet}  %Required
\usepackage{courier}  %Required
\usepackage{url}  %Required
\usepackage{graphicx}  %Required
\begin{document}
\title{Integrative Biological Simulation, Neuropsychology, and AI Safety}
\author{Gopal P. Sarma$^{1}\thanks{Email: gopal.sarma@emory.edu}$, Adam Safron$^{2}$, and Nick J. Hay$^{3}\thanks{The views expressed herein are those of the author and do
not necessarily reflect the views of Vicarious AI.}$\\
School of Medicine, Emory University, Atlanta, GA USA \\
Department of Psychology, Northwestern University, Evanston IL USA \\
Vicarious AI, San Francisco, CA USA \\
}
\maketitle
\begin{abstract}
We describe a biologically-inspired research agenda with parallel tracks aimed at AI and AI safety.  
The bottom-up component consists of building a sequence of biophysically realistic simulations of simple organisms 
such as the nematode \emph{Caenorhabditis elegans}, the fruit fly \emph{Drosophila melanogaster}, and the 
zebrafish \emph{Danio rerio} to serve as platforms for research into 
AI algorithms and system architectures.  The top-down component consists of an approach to value alignment that grounds
AI goal structures in neuropsychology, broadly considered.  Our belief is that parallel pursuit of these tracks will inform 
the development of value-aligned AI systems that have been inspired by embodied organisms with sensorimotor integration. 
An important set of side benefits is that the research trajectories we describe here are grounded in 
long-standing intellectual traditions within existing research communities and funding structures. In addition, these research 
programs overlap with significant contemporary themes in the biological and psychological sciences such as 
data/model integration and reproducibility.
\end{abstract}

\frenchspacing
\section{Introduction}
Bostrom's orthogonality thesis states, under certain weak assumptions, that the intelligence of an agent 
and its goal structure are independent variables \cite{bostrom2014superintelligence}.  The orthogonality 
thesis is a useful conceptual tool for correcting for anthropomorphic bias, i.e. the assumption that an arbitrary 
agent will behave in a manner similar to human beings.  Particularly in a climate of fear and uncertainty about future AI systems, 
the orthogonality thesis can be a helpful framing to encourage reasoning more clearly about the risks of advanced AI systems 
and to dislodge concerns arising from science fiction movies and sloppy journalism. However, 
from an engineering standpoint, it is worth considering that the design of safe, 
superintelligent AI systems may \emph{benefit} from examining system architectures in which the intelligent 
substrate and fundamental goal structure of an agent have been \emph{intentionally} coupled.  
We have used the phrase \emph{anthropomorphic design} 
to refer to approaches in which AI systems are built to possess commonalities with human 
neuropsychology \cite{sarma2017mammalian,sarma2018ai,sotala2016defining}. \\

Anthropomorphic design is best described in reference to the concept of \emph{value alignment}, which
refers to the construction of AI systems that act in accordance with human values throughout their operation \cite{bostrom2014superintelligence,russell2016should}.
In recent years, value alignment has been decomposed into specific sub-problems representing
possible failure modes of AI systems \cite{amodei2016concrete}.  For instance, consider the sub-problem of ``avoiding negative side effects.'' Often times
when we specify a goal, we implicitly assume additional criteria from our model of the world and our value systems which go
unstated \cite{shanahan2006frame}.  As an example, if we have a fully autonomous waste management system to which we give the goal ``reduce pollution,"
we do not want the system to go dump all of the pollution in a neighboring region outside of the range of its sensors, thereby
``reducing pollution.'' A more robust understanding of how to design
systems which accomplish their objective while minimizing external impact may allow us to tackle an essential component of value alignment.
In a similar vein, ``avoiding reward hacking,'' 
``scalable oversight," ``safe exploration,'' and ``robustness to distributional shift'' are all specific properties researchers
have identified that should be possessed by value-aligned AI systems \cite{amodei2016concrete}.  
Decomposing value alignment into more basic
constituents carries the additional benefit that \emph{test suites} can be designed to verify 
that agents possess those properties in simulated environments \cite{leike2017ai}. Although AI systems today are not powerful enough
for value \emph{mis}-alignment to have particularly negative consequences, the simple frameworks researchers have designed
to date are a promising starting point to evolve into substantially more intricate environments for ensuring the safety of future AI systems.  \\

We have previously argued that research in affective neuroscience and related disciplines aimed at 
grounding human values in neuropsychology may
provide important conceptual foundations for understanding value alignment in AI systems. We use neuropsychology in the broadest
possible sense of referring to efforts aimed at correlating psychological function with underlying neural architecture. There are several practical benefits
that might emerge from such a research program, i.e. one aimed at anthropomorphic design.  Having more detailed prior information about human values may
allow a sophisticated AI system to learn from fewer examples. Similarly, it may enable practical implementations
of AI safety techniques that would otherwise be computationally intractable \cite{sarma2017mammalian,sarma2018ai}.  \\

In this brief position paper, we describe a bottom-up approach to understanding AI architectures that dovetails with both the neuropsychology and test suite based approaches to ensuring value alignment described above. This program is aimed at building realistic biophysical simulations of simple nervous systems 
which incorporate biomechanics in a simulated environment. Because of key architectural commonalities in the brain plans of vertebrates, 
and even some invertebrates, we believe that this research is a natural complement 
to the neuropsychology-based approach to value alignment we have described previously.
Our strong intuition is that parallel pursuit of these goals will not only lead to fundamental advances in AI algorithms, but also architectural
insights into ensuring value alignment.  \\

The 
contributions of this manuscript are two-fold: \emph{(i)} We introduce a set of research
objectives in neuroscience that are well positioned to give rise to significant advances in AI and which have received
little attention by the AI safety community.  \emph{(ii)} We suggest two existing approaches to AI safety that may integrate with this research paradigm:
a neuropsychology-based approach to value alignment and test suites for agent-based AI systems in simulated environments.  We emphasize at the outset that the perspective we take here is largely \emph{descriptive.}  Although we have provided a novel framing, much of the research we discuss here is actively ongoing in a diverse set of communities in the biological
sciences.  Our objective in this position paper, therefore, is to call attention to a potential path to powerful AI systems which can be entirely justified on their intrinsic value for the biological
sciences and which has received little attention by the safety community. It is essential, therefore, as this research progresses at an increasing pace, for parallel steps to be taken to ensure the safety of the resulting
systems.  

\section{Integrative Biological Simulation}
\emph{\textbf{Claim 1: Simple organisms show complex behavior that continues to be difficult for modern AI systems. Neuronal simulations in virtual environments will allow these biological architectures to be used for AI research.}} \\

Integrative biological simulations refer to computational platforms in which diverse, process-specific models, often operating
at different scales, are combined into a global, composite model \cite{sarma2017integrative}.  Examples include OpenWorm, an internationally coordinated open-science project working towards a realistic biophysical simulation of the nematode \emph{Caenorhabditis elegans}, 
Neurokernel, a project with some parallels to OpenWorm aimed at simulating \emph{Drosophila melanogaster}, Virtual Lamprey, 
a computational platform for understanding vision and locomotion in the lamprey, BlueBrain, an effort
to build a detailed model of the rat cortical micro-column, and the Human Brain Project (HBP), an ambitious successor to BlueBrain which aims
to extend this platform to an entire human cerebral cortex \cite{sarma2018openworm,givon2016neurokernel,lamprey,markram2015reconstruction,amunts2016human}.
Such platforms may serve as points of integration for data and computational models. The result is a shared structure that can be used
by an entire community of researchers to test novel hypotheses, create a tighter feedback loop between experimental and theoretical
research, and ensure the reproducibility and robustness of the underlying research output. \\

In the AI community, awareness of these research programs has primarily been informed by the efforts of BlueBrain and HBP 
to simulate large regions of mammalian cortical tissue.  We are sympathetic in many ways to the aims of these projects. 
However, we believe that an under-appreciated set of approaches complementing their work consists of
using analogous software infrastructure to develop simulations of organisms far below the complexity of mammals or vertebrates. 
\emph{C. elegans}, with only 302 neurons, shows simple behavior of learning and memory. \emph{Drosophila melanogaster}, despite only
having 10$^{5}$ neurons and no comparable structure to a cerebral cortex, has sophisticated spatial navigation abilities easily rivaling
the best autonomous vehicles with a minuscule fraction of the power consumption.  The zebrafish \emph{Danio rerio} has on the order of 10$^{7}$
neurons and has been a model system in neuroscience for several decades.  Moreover, recent efforts to perform whole-brain functional imaging
in the larval zebrafish may make this a particularly attractive target for future integrative simulation platforms \cite{ahrens2013whole}.  
Although much of this research has been motivated by neuroscientific
aims and connections to the study of human disease processes, the implications for AI research are significant.  Well-engineered
software platforms which allow for rapid iteration on existing architectures without the constraints of biological realism will allow
AI researchers to test novel hypotheses in embodied organisms in simulated environments [see related work in the animat community \cite{strannegaard2017animat,Wilson91theanimat}].  Real-time visualization of nervous system activity will allow for a deeper understanding of how AI algorithms such as backpropagation, belief propagation, or reinforcement learning may approximate what is observed in nature.  \\

Coupling nervous system activity to drive a simulated body is a tractable approach with organisms such as \emph{C. elegans} and \emph{Drosophila}.
In OpenWorm, for example, the Boyle-Cohen model of neuromuscular coupling allows for the output
of connectome dynamics to drive the activation of body wall muscles and a simulated body \cite{boyle2008caenorhabditis,gleeson2018c302,palyanov2018three}.  
Similar models are likely achievable with \emph{Drosophila} as well. Indeed, the Neurorobotics Platform of HBP is working towards
a general platform for interfacing realistic neural network simulations with robotic bodies \cite{HBP_neurorobotics,oberts2016brain}. 
The incorporation of biomechanics into 
these simulations can be justified on biological grounds.  For instance, understanding the effects of 
anti-psychotic or anti-epileptic medications in model organisms is simplified if researchers can observe changes 
in behavioral patterns, rather than having to interpret
high-dimensional data streams of neuronal activity. However, there are reasons to think that sensorimotor integration may be particularly
valuable from a purely AI perspective. \\

As others have argued, despite the significant advances arising from the use of deep representations in neural networks, current 
AI systems continue to lack many of the qualities of fluid intelligence observed in human beings, particularly in the ability to learn
concepts from a relatively small number of examples.  One hypothesis is that, unlike modern deep learning systems, human concepts 
are grounded in rich sensorimotor experience. Despite significant work in transfer learning and domain adaptation, modern systems
are largely restricted in their domain of application.  The lack of behavior-based concept representation may be a limiting
factor in current state-of-the-art systems \cite{hay2018behavior,krichmar2018neurorobotics,HBP_neurorobotics}.  
Simulations of simple, embodied organisms with realistic virtual environments may provide
platforms for AI research aimed at understanding the interplay between concept representation and embodiment. 
Moreover, used in a modular or hierarchical fashion, contemporary techniques such a deep learning may prove to be powerful components of future
iterations of these platforms. 

\section{Neuropsychology and Value Alignment}
\emph{\textbf{Claim 2: Value-alignment research may benefit from insights in neuropsychology and comparative neuroanatomy.}}\\

We have argued previously for an approach to value alignment which grounds an understanding of human values in neuropsychology \cite{sarma2017mammalian,sarma2018ai}.
In this section, we reproduce the broad outlines of this framework before discussing how these parallel research tracks may 
come to intersect. Our approach is loosely based on research in affective neuroscience, which aims to categorize
emotional universals in the mammalian class and correlate them with an underlying neurological substrate \cite{panksepp1998affective}.  
We use a broad interpretation of the term neuropsychology to denote research aimed at correlating psychological behavior with underlying
neural architecture; other related and possibly relevant fields of 
research include contemplative neuroscience, neuropsychoanalysis, biological anthropology, and 
comparative neuroanatomy, to name just a few. \\

It is possible that values and motivations are fundamentally grounded in emotions for human beings. 
If our emotional substrate is shared with other mammals, or even more broadly with other vertebrates and animals, 
it suggests that our value systems can be decomposed in ways that inform neuroscience-based AI architectures.  For example, one 
possible (non-exclusive) decomposition of human values is the following:
\begin{enumerate}
\item \textbf{Internal reward systems shared by all mammals}: In the taxonomy of affective neuroscience, these include 
play, panic/grief, fear, rage, seeking, lust, and care. This may also include curiosity and the acquisition of skills.
\item \textbf{Internal reward systems with human-specific elaborations}: For example, uniquely
human social behaviors such as family membership, group affiliation, story telling, and gift giving. 
\item \textbf{Products of human deliberation/cognition on our values:} The many complex features of value systems 
produced by several millennia of human social and cultural evolution; likely mediated by cultural inheritance. 
\end{enumerate}

An alternate version which we have previously suggested is to view human values as consisting of \emph{1) mammalian values 2) human cognition and 3) several millennia of human social and cultural evolution} \cite{sarma2017mammalian}. Decompositions such as these might allow AI systems to begin with a more nuanced understanding of human values that is
then refined over time through observation, hypothesis generation, and human interaction. 
For an agent that is actively interacting with the world during the learning process, a more informative prior may allow
a system to learn from fewer examples, directly translating into a reduced risk of adverse outcomes. 
Likewise, consider that our values and culture are instilled in children by selective exposure to carefully chosen environments. 
A neuropsychological understanding of human values may allow us to make similarly strategic choices for 
AI systems in order to minimize the time required to achieve strong guarantees of value alignment 
\cite{evans2016learning,christiano2017deep}. Moreover, systems with human-inspired architectures may lead to
natural avenues for addressing issues of transparency and intelligibility of AI decision making \cite{bryson,floridi}.

\section{Synthesis}
\textbf{\emph{Claim 3: Significant synergy may be achieved by coupling the two research programs described above.}} \\

Thus far, we have discussed organisms which lie very far apart on the evolutionary tree. \emph{C. elegans} and 
\emph{Drosophila} possess only 10$^{2}$ - 10$^{5}$ neurons and the zebrafish \emph{Danio rerio} roughly 10$^{7}$ neurons, whereas mammalian brains range from 
10$^{8}$ neurons in the brown rat to 10$^{10}$ neurons in human neocortex. However, by the time
we reach \emph{Drosophila} we are already confronting a brain with many high-level architectural features which higher animals share, 
such as two lobes and distinct functional processing regions. Moreover, insects share many neurochemical motivational systems with 
vertebrates and even higher mammals \cite{panksepp1998affective}.
Proceeding up the evolutionary tree a little further, sophisticated 
brain centers involved in motor coordination, such as the basal ganglia, are known to be conserved across vertebrates (including
zebrafish), and may have homologous 
structures in arthropods \cite{grillner2016basal}.  In other words, viewed as platforms for research into value-aligned AI systems, there may be clues
even from invertebrates and simple vertebrates for how the insights from top-down, neuropsychology-based approaches may be used 
to design AI systems that possess far greater levels of transparency, intelligibility, and goal structure stability than
we see in nature or in our current AI technologies. We believe that a healthy level of interaction between the otherwise disparate
communities pursuing these lines of research is the most fruitful way to uncover such clues and to establish clear research directions 
which lie at the intersection of the two approaches. Moreover, BlueBrain/HBP are already tackling the substantially more difficult
challenge of simulating mammalian brains. The success
of these projects will only complement insights that arise from approaches oriented towards simulation of simple organisms.  \\

Another point of intersection between integrative biological simulation and current research
in AI safety is to extend the concept of test suites for
RL agents to the virtual environments of simulated
organisms \cite{leike2017ai}.  As we discussed above, test suites have emerged in the AI safety community as a way to 
operationalize value alignment into practical, albeit long-term, development strategies for AI systems.  By decomposing value alignment into specific sub-problems, simulated environments can be created to assess the degree to which artificial agents solve specific tasks while
adhering to global safety constraints.
Problems such as safe interruptibility, avoiding negative side effects, reward gaming, distributional shift, and others should be adaptable to virtual biological organisms. For example, to what degree do we see variation in susceptibility to reward hacking 
(i.e. addictive behaviors) in the animal kingdom? Lifting biological constraints, can we augment simulated architectures with modules
to reduce the risk of such behavior?

\section{Discussion and Future Directions}
We reiterate that the perspective taken here is largely a descriptive one, as many of the topics we have discussed are actively
being pursued by researchers in the biological and psychological sciences.  The novel contribution of this manuscript is to frame
this research in the context of AI safety. Therefore, in arguing that these two research agendas 
be ``coupled,'' our intent is to promote community interaction and not necessarily dual-pronged approaches within individual research groups.  Given the relative
infancy of these ideas, we suspect that much discussion will be necessary before identifying concrete points of overlap. \\

For the integrative simulation projects, we encourage interested researchers to consult the publications of the respective research groups to find
concrete points of entry. For those attracted to expanding
the repertoire of simple organisms that have such platforms, there are many commonalities in the necessary software infrastructure, 
with tools such as NEURON for simulating Hodgkin-Huxley type models, BluePyOpt for 
extracting kinetic parameters for experimental data, and NetPyNE/Bionet for specifying 
network models \cite{hines1997neuron,van2016bluepyopt,gratiy2018bionet}.  Aside from the connectome, an area where
there are relevant differences between these organisms is in the gene expression of ion channels. Efforts such as ChannelPedia, NeuroMLDB, ModelDB, and Open Source Brain, all share the goal of enabling storage and re-use of neuroscience data and models 
\cite{ranjan2011channelpedia,gleeson2012open}.  
Expanding the scope of these resources
to include ion channel data and models for a variety of species would be a key enabler of this research agenda. We suspect 
that there is literature on comparative neuroanatomy that will give us insights into 
promising directions to pursue on the lower part of the evolutionary tree.   \\

With regards to the top-down approach to value-alignment, as we emphasized in our previous manuscript, a key obstacle 
is the widespread concern of reproducibility issues in the biological and psychological literature \cite{sarma2018ai,sarma2017doing}.  
Therefore, we are of the conviction that the most immediate next step is to create a community-driven replication effort aimed at developing a more
robust body of knowledge with which to base future research.  To that end, we have created a project using the Open Science Framework
where we are currently collecting suggestions for candidate studies which would be of high value to either directly replicate or validate through some other 
means.\footnote{\url{https://tinyurl.com/AI-reproducibility}} We are particularly interested in using iterated expert elicitation 
methods such as RAND Corporation's Delphi protocol to encourage
consensus building among researchers \cite{brown1968delphi}.  \\

Finally, regarding the development of test suites for simulated organisms, we have no illusions as to the 
difficulty of the challenge.  Understanding how to translate the highly simplified models of current AI safety frameworks to the complex
neural networks of real organisms in realistic physical environments will be a substantial undertaking.  However, we believe that such a synthesis is both 
necessary and desirable, as it may provide insight into hybrid approaches which take advantage of both modern AI and 
simulated biology to build sophisticated value-aligned systems.  

\section*{Acknowledgments}
We would like to thank Owain Evans, Tom Everitt, and several anonymous reviewers for insightful discussions and feedback on the manuscript.  

\fontsize{9.5pt}{10.5pt} 
\selectfont
\bibliographystyle{aaai}
\bibliography{simulation_neuropsychology}
\end{document}